\documentclass[conference]{IEEEtran}
\IEEEoverridecommandlockouts
\usepackage{cite}
\usepackage{amsmath,amssymb,amsfonts}
\usepackage{algorithmic}
\usepackage{graphicx}
\usepackage{tabularx}
\usepackage{textcomp}
\usepackage{xcolor}
\usepackage{comment}
\def\BibTeX{{\rm B\kern-.05em{\sc i\kern-.025em b}\kern-.08em
    T\kern-.1667em\lower.7ex\hbox{E}\kern-.125emX}}
    
\usepackage{array,booktabs}

\usepackage{array}
\usepackage{xcolor}
\newcolumntype{P}[1]{>{\centering\arraybackslash}p{#1}}
\newcolumntype{M}[1]{>{\centering\arraybackslash}m{#1}}
\usepackage{multirow}
\usepackage{multicol} 
\usepackage[normalem]{ulem}
\definecolor{cg_green}{RGB}{0, 110, 44}

\title{Augmentation Methods on Monophonic Audio for Instrument Classification in Polyphonic Music\\
\thanks{$^*$The first two authors contributed equally.}
}

\begin{document}

\author{\IEEEauthorblockN{Agelos Kratimenos$^*$}
\IEEEauthorblockA{\textit{School of ECE, NTUA,} Athens, Greece\\
ageloskrat@yahoo.gr}
\and
\IEEEauthorblockN{Kleanthis Avramidis$^*$}
\IEEEauthorblockA{\textit{School of ECE, NTUA,} Athens, Greece \\
kle.avramidis@gmail.com}
\and 
\IEEEauthorblockN{Christos Garoufis}
\IEEEauthorblockA{\textit{School of ECE, NTUA,} Athens, Greece\\
cgaroufis@mail.ntua.gr}
\and
\hspace{3cm}
\IEEEauthorblockN{Athanasia Zlatintsi}
\IEEEauthorblockA{ \hspace{3cm} \textit{School of ECE, NTUA,} Athens, Greece \\
\hspace{3cm} nzlat@cs.ntua.gr}
\and
\hspace{-2.5cm}
\IEEEauthorblockN{Petros Maragos}
\IEEEauthorblockA{\hspace{-2.5cm} \textit{School of ECE, NTUA}, Athens, Greece \\
\hspace{-2.5cm} maragos@cs.ntua.gr}
}

\maketitle

\begin{abstract}
Instrument classification is one of the fields in Music Information Retrieval (MIR) that has attracted a lot of research interest. However, the majority of that is dealing with monophonic music, while efforts on polyphonic material mainly focus on predominant instrument recognition. In this paper, we propose an approach for instrument classification in polyphonic music from purely monophonic data, that involves performing data augmentation by mixing different audio segments. A variety of data augmentation techniques focusing on different sonic aspects, such as overlaying audio segments of the same genre, as well as pitch and tempo-based synchronization, are explored. We utilize Convolutional Neural Networks for the classification task, comparing shallow to deep network architectures. We further investigate the usage of a combination of the above classifiers, each trained on a single augmented dataset.  An ensemble of VGG-like classifiers, trained on non-augmented, pitch-synchronized, tempo-synchronized and genre-similar excerpts, respectively, yields the best results, achieving slightly above 80\% in terms of label ranking average precision (LRAP) in the IRMAS test set.
\end{abstract}

\begin{IEEEkeywords}
instrument classification, audio mixing, data augmentation, deep learning, ensemble learning
\end{IEEEkeywords}

\section{Introduction}
\label{sec:intro}
With the term “music” we refer to interesting combinations of one or more instrument sounds and sometimes vocals. Humans are able, due to their natural ability, to identify these sounds when they listen to music. 
Of course, this ability gets harder depending on the number of instruments, the performance style and also when perceptually similar instruments play simultaneously. It is easy to infer that for computers and computing algorithms this task gets even more complex, which has led to increased research activity in various fields of MIR. \par

\begin{figure}[t]

\begin{minipage}[t]{1.0\linewidth}
  \centering
  \centerline{\includegraphics[width=9cm]{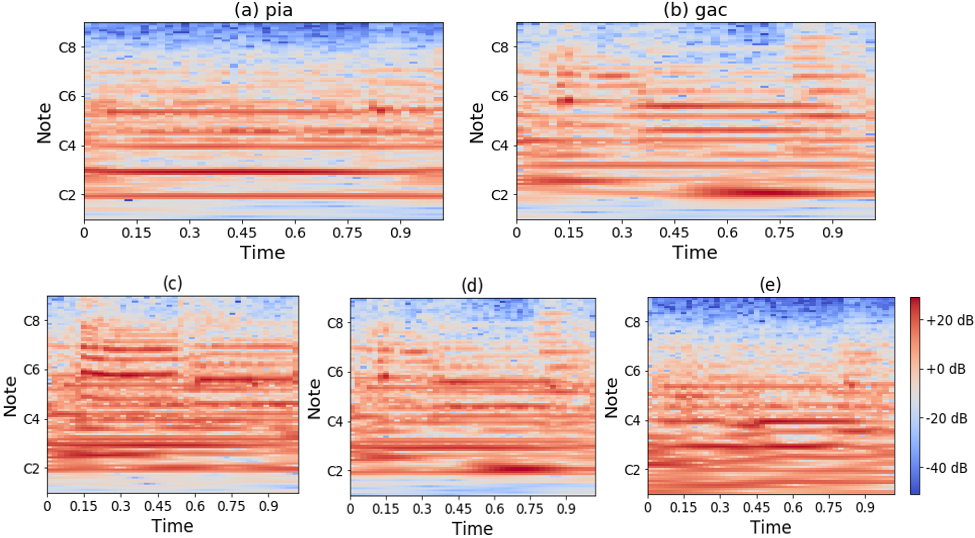}}
\end{minipage}

\label{sec:foot}
\caption{CQT spectrograms of 2 segments derived from piano and guitar. (a) Piano, (b) Acoustic Guitar, (c) Random Mix, (d) Tempo Mix, (e) Pitch Mix}
\vspace{-0.8cm}
\end{figure}

Instrument Classification (IC) on monophonic data has already been successful. On the other hand, the equivalent polyphonic task is much harder, because of the existence of similar instruments and the superposition of time-frequency features, such as pitch and timbre, from different instruments. However, this task is being thoroughly researched, as it can lead to significant 
advances and applications. The ability to determine which instruments are playing at each time can provide useful insight into musical structure and therefore can act in assistance to a variety of fundamental MIR tasks, such as music browsing \cite{b1}, auto-tagging \cite{b2}, music automatic transcription \cite{b3} and source separation \cite{b4}.\par
An important research constraint for instrument classification is the lack of easily accessible data, since it is quite hard to create a large dataset of polyphonic audio that is correctly annotated. On the other hand, monophonic data can be easily accessed, collected, and automatically annotated. Inspired by this, in this paper we investigate how to efficiently augment monophonic data in order to train an instrument classifier for polyphonic music, thus focusing on mixing methods. To the best of our knowledge, this is the first study that focuses on mixing monophonic audio as a method to create a dataset suitable for polyphonic music tasks. As mixing methods, we investigate superimposition between perceptually similar audio segments, as well as superimposition of audio segments after alignment on the temporal or frequency scale. A secondary
contribution of this work is a comparative study between the architecture proposed in \cite{b15} for the task of IC and deeper, VGG-like \cite{b25} architectures, such as the one used in \cite{b21}, indicating that a deeper architecture is necessary to integrate the required features for successful instrument classification. 
\par The rest of this paper is organized as follows: Section II provides a review of related work in the recent literature. The proposed augmentation methods, used to convert our initial monophonic dataset into a polyphonic one, are analyzed in Section III. 
Section IV deals with the experimental setup, describing the pre-processing and the model architectures we exploit, while the evaluation metrics utilized are also introduced. In Section V we present the results of our experiments and evaluate the augmentation strategies and finally, in Section VI we conclude our remarks and propose plausible future directions.

\section{Related Work}
\label{sec:format}

\subsection{Instrument Classification}

Traditionally, research work using monophonic audio focused mainly on recorded isolated notes, using for instance cepstral coefficients and temporal features \cite{b5} or phase information with MFCCs \cite{b6}. For an overview on note-level instrument recognition, readers are referred to \cite{b7}.

\par More recent works however deal with polyphonic sounds, which are closer to actual music. Until very recently, the vast majority of works on instrument classification used datasets of solo recordings or excerpt-level annotations (e.g. IRMAS \cite{b11}). A number of research studies tackled this problem by using synthesized polyphonic audio from recorded single tones \cite{b8} or MIDI files \cite{b14,b32}. 
In \cite{b9} hand-crafted audio features along with dimensionality reduction techniques were used in multi-instrument segments. However, artificially produced polyphonic music is still far from professionally produced music. Factors like timbre, style, or even the recording quality affect the recognition performance. 
\par  Since labeling the presence of instruments for each frame requires a lot of effort and time, few projects deal with frame-level analysis and datasets with frame-level instrument labels have only emerged over the recent few years \cite{b12,b13}. Among state-of-the-art efforts, Hung et al. \cite{b14,b32} exploit musical traits like timbre and pitch to make frame-level predictions using MusicNet \cite{b13}. In \cite{b15}, Gururani et al. experiment with various temporal resolutions in the classification task, while in \cite{b26}, the usage of an attention layer was shown to improve classification results when applied to a set of features extracted from a pre-trained VGG net. \par

\subsection{Data Augmentation Techniques on Speech and Music}

  Data augmentation has generally been reported to prevent overfitting issues, something that has resulted in numerous related experiments in music recognition recently. Frequently used data augmentation techniques in MIR and speech recognition include addition of noise \cite{b27,b29,b30}, simulation of spatial effects \cite{b27}, pitch shifting \cite{b28,b29}, time stretching \cite{b28,b29}, and amplitude shifiting \cite{b29}\footnote{Open-source packages that support a number of these transformations include muda \cite{b16} and rubberband (https://breakfastquay.com/ rubberband)}, while in \cite{b17} an augmentation policy that includes feature warping and masking on spectrograms has provided state-of-the-art results. The work that is closest to ours, with regards to the type of augmentation performed, is that of Lee et al. \cite{b31}, where vocal and instrumental audio segments are mixed together by virtue of their similarity with regards to tempo, beat and key. 

\begin{figure}[t]

\begin{minipage}[t]{1.0\linewidth}
  \centering
  \centerline{\includegraphics[width=9cm]{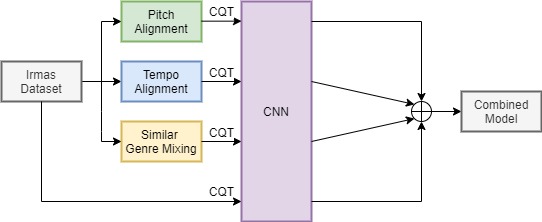}}
\end{minipage}

\label{sec:foot}
\caption{Proposed pipeline. The best model occurs from an aggregation of the individual models produced by each mixing method.\newline}
\vspace{-0.6cm}
\end{figure}  
  
\section{Methodology}
\label{sec:majhead}

\subsection{Initial Dataset}
\label{ssec:subhead}

The dataset used in order to train our models is the IRMAS dataset \cite{b11}, which is divided into a training and a testing set. We use the training set, which consists of 6705 3-sec audio snippets at 44.1kHz, annotated with one of the 11 available predominant instrument labels: piano (pia), acoustic guitar (gac), electric guitar (gel), violin (vio), cello (cel), saxophone (sax), clarinet (cla), trumpet (tru), human voice (voi), organ (org) and flute (flu). We choose to cut each track into 1-sec segments, since this temporal resolution yielded the optimal results in \cite{b21}. Our experiments show that the imbalance between certain classes is not an important factor during classification, so we do not consider further balancing.

\subsection{Augmentation Methods}
\label{ssec:subhead}
We apply a number of different mixing methods to our 1-sec dataset, in order to make it polyphonic and larger in quantity, thus capable of yielding better estimations based on the same monophonic data. We consider a naive random mixing method as well as mixings with synchronizations on 3 different audio features: pitch, tempo and genre. These features are chosen because they encapsulate most of the meaningful information of an audio track. Moreover, we expect that by synchronizing audio tracks over certain features, a classifier would be able to isolate better other meaningful attributes from an audio mixture. In this paper we experiment with mixes of pairs of tracks from the 11 classes, which adds up to 55 additional classes. The mixes are implemented using the PyDub library.

\subsubsection{Random Mixing}
\label{sssec:subsubhead}
For each combination of instruments we define the class with the fewer audio segments, shuffle the samples of both sets and simply overlay them one by one. As a result, we produce a new dataset of mixed tracks without any repetition. Thus, every new dataset will have as many tracks as the smallest of the 2 monophonic ones. 

\subsubsection{Pitch-Sync Mixing}
\label{sssec:subsubhead}
The intuition behind the specific augmentation strategy is that, by pitch alignment of the overlapped segments, we force our network to focus on specific spectral and timbral characteristics. In order to align two audio segments, we first use the CREPE pitch prediction model \cite{b22}, which estimates the pitch of an audio signal using 10~msec windows. After acquiring the main frequencies $f_1$ and $f_2$ of the two segments, we calculate the required frequency shift per segment, in semitones $s=12\log_2({f_1} / {f_2})$.%
We then smooth the frequency shift vector. Initially, we apply a median filter of kernel size $9$, covering 90ms, to the initial vector. Afterwards, we merge the neighboring segments of the same or similar ($\pm 1$) semitone shifts, as well as those with duration less than 70ms with their preceding segments. 
Finally, we shift the second segment by the computed semitone shifts using pyrubberband. The mixing of a pair of such tracks is done following the procedure described above, resulting in 55 sets of mixed tracks without repetition.

\begin{table}[t]
    \centering
    \caption{Initial and Proposed Model Architecture. Depth: $d_i=64, 128, 256, 640$, Max Pooling: $p_i=(2,2), (2,2),(3,3),$ $(3,3)$}
    \begin{tabular}{|c|c|c|}
    \hline
    & \textbf{Initial 1-Conv Model \cite{b15}}  & \textbf{Proposed 2-Conv Model} \\ \hline
    \multirow{4}{*}{$\times4$} & Conv2D $(3\times3,d_i)$ &  $2\times$Conv2D $(3\times3,d_i)$ \\\cline{2-3}
     & \multicolumn{2}{c|}{Batch Normalization}               \\ \cline{2-3}
     & ELU             & LeakyReLU $(a=0.3)$          \\ \cline{2-3}
     & \multicolumn{2}{c|}{Max Pooling $(p_i)$}      \\ \cline{2-3}
     & \multicolumn{2}{c|}{Dropout $(0.2)$}                    \\ \hline
     & \multicolumn{2}{c|}{Dense $(1024)$}            \\ \hline
     & ELU             & LeakyReLU $(a=0.3)$        \\ \hline
     &\multicolumn{2}{c|}{Batch Normalization}   \\ \hline
     & \multicolumn{2}{c|}{Dropout $(0.5)$}                 \\ \hline
     & \multicolumn{2}{c|}{Dense $(11)$}         \\ \hline
     &  \multicolumn{2}{c|}{Sigmoid Activation}    \\ \hline
    \end{tabular}
    \label{tab:sometab}
    \vspace{-0.4cm}
\end{table}

\subsubsection{Tempo-Sync Mixing}
\label{sssec:subsubhead}
We proceed in tempo-syncing since instruments often play in the same beat when they co-perform. For this method, we need to calculate the tempo, in Beats per Minute (BPM), for each audio track pair that will be overlaid. The 1-sec resolution however is not adequate for BPM detection, so we exploit the initial 3-sec IRMAS tracks and cut them into 1-sec segments after mixing. For a pair of tracks to be mixed, we detect their BPM using Librosa \cite{b20} and we apply time stretching using pyrubberband. Since the duration of the stretched segment will change, we either cut the segment or repeat its beginning, in order to set its duration to 3 sec. The final tracks are mixed with PyDub and cut afterwards.

\subsubsection{Similar Genre Mixing}
\label{sssec:subsubhead}
As a final mixing method, we combine segments that belong to the same genre. IRMAS proves to be ideal for this method, since the data are already annotated with genre labels. The available categories are classic, pop/rock, jazz/blues and country/folk. We produce the mixed tracks as in Random Mixing, by matching only segments from the same genre category. By doing so, we expect to better approximate actual music tracks, each of which relates to a specific genre. 

\section{Experimental Setup}
\label{sec:majhead}

\subsection{Pre-Processing}
\label{ssec:subhead}
We produce a new set of music segments by applying each of the above augmentation methods. For each set, every track is downsampled to 22.05 kHz, downmixed to mono and also normalized by the root mean square energy. We experiment with Constant Q Transform (CQT) as a feature representation of the audio segments, as it has been proven adequate for various MIR tasks, serving a perceptually motivated frequency scale \cite{b23}. Each segment is transformed into a CQT-spectrogram, using $96$ bins, $12$ bins per octave and $256$ hop length, resulting in a $96\times87$ matrix. \par The set of spectrograms is then partitioned into 5 subsets, each containing an equal amount of samples from every class. We use this scheme in order to perform multiple training sessions, using each subset in rotation as validation data and the remaining 4 for fitting the classifier. The advantage of this training method is that every sample is exploited for both training and validation, thus resulting in more representative estimations and further statistical inference.

\begin{table}[] \centering
\caption{Comparison of Model Architectures} \label{tab:sometab}

\begin{tabular}{c|c|c|c}
Mixing Dataset         & Model  & LRAP   & AUC \\ \hline \hline
\multirow{2}{*}{Monophonic}      &  Gururani et al. {[14]} & 0.738          & 0.839          \\ 
                                 & Proposed        & \textbf{0.767} & \textbf{0.843} \\ \hline
\multirow{2}{*}{Random}          & Gururani et al. {[14]} & 0.750          & 0.853       \\ 
                                 & Proposed        & \textbf{0.776} & \textbf{0.861}  \\ 
\end{tabular}
\vspace{-0.4cm}
\end{table}

\subsection{Model Architecture and Training}
\label{ssec:subhead}
As a baseline model we use the network proposed by Gururani et al. \cite{b15}, which consists of four convolutional layers and two dense layers. Every convolutional layer uses zero padding and is followed by Batch Normalization, Max Pooling and Dropout layers. ELU activation is used throughout the model and a sigmoid activation is chosen for the output layer, as it gives a probability distribution around all predictable classes. We further experiment with the architecture and we propose a model that uses double layers at each convolutional block of the model, while we also replace the ELU activation function with a leaky-ReLU \cite{b18}. The initial and proposed architecture can be seen in Table I.

\par We train both networks using binary cross-entropy loss. Adam optimizer is used to optimize the loss function, with an initial learning rate of 0.0001 and 10\% decay rate per epoch, while the batch size is set to 128. We perform Learning Rate Reduction and Early Stopping by monitoring the validation loss with a patience of 5 and 7 epochs respectively.

\begin{table*}[t]
    \centering
     \caption{Results for each augmentation strategy compared to the initial dataset, using the proposed architecture} \label{tab:sometab}
    
    \begin{tabular}{|c|P{1.8cm}|P{1.8cm}|P{1.8cm}|P{1.8cm}|P{1.8cm}|P{1.9cm}|} \hline
    
    \textbf{Metrics\textbackslash  Mix Methods} & Monophonic &
    Random &
    Genre &
    Tempo & 
    Pitch & 
    Combined \\ \hline

    LRAP&
    0.767 $\pm$ 0.008  &
    0.776 $\pm$ 0.006  & 
    0.774 $\pm$ 0.005  & 
    0.790 $\pm$ 0.003  & 
    0.795 $\pm$ 0.002  & 
    \textbf{0.805 $\pm$ 0.002}  \\
    
    Mean AUC & 
    0.843 $\pm$ 0.002  & 
    0.861 $\pm$ 0.001  & 
    0.862 $\pm$ 0.003  & 
    0.865 $\pm$ 0.002  & 
    0.856 $\pm$ 0.001  & 
    \textbf{0.874 $\pm$ 0.001}  \\
    
    $F_1$ micro & 
    0.616 $\pm$ 0.009  & 
    0.624 $\pm$ 0.006  & 
    0.617 $\pm$ 0.004  & 
    0.628 $\pm$ 0.003  & 
    0.635 $\pm$ 0.002  & 
    \textbf{0.647 $\pm$ 0.003}  \\
    
    $F_1$ macro & 
    0.506 $\pm$ 0.006 & 
    0.528 $\pm$ 0.006  & 
    0.519 $\pm$ 0.004  & 
    0.531 $\pm$ 0.003  & 
    0.532 $\pm$ 0.003 & 
    \textbf{0.546 $\pm$ 0.004}  \\ \hline
    
    \end{tabular}
    \vspace{-0.4cm}
\end{table*}

\subsection{Evaluation Metrics}
\label{ssec:subhead}

We evaluate the proposed models at the IRMAS test set, consisting of 2355 music tracks. Since our models are trained to predict the instruments in 1-sec segments and the duration of the testing track ranges from 5 to 20 sec, we average the per-frame predictions and extract a single total prediction for each track. While this would normally result in flawed predictions, the fact that each labeled instrument is active for the whole duration of the track allows the above manipulation. In the following, we utilize the next three metrics:

\subsubsection{Label Ranking Average Precision (LRAP)}
\label{sssec:subsubhead}
The LRAP metric has been proposed in \cite{b24} and is suitable for multi-label classification evaluation. LRAP is threshold-independent for recognition tasks and measures the classifier's ability to assign higher scores to the correct labels associated to each sample, compared to the false ones. Formally, given $n$ test samples, a binary-indicator matrix $y$ of their ground truth labels and the score $f$ of each label, LRAP is defined as:

\vspace{-0.4cm}
$$\mathrm{LRAP}(y,f) = \frac{1}{n}\sum_{i=0}^{n-1}{\frac{1}{\|y_i\|_0}} \sum_{j:y_{ij}=1}{\frac{\ell_{ij}}{\mathrm{rank}_{ij}}},$$
\noindent where we define $\ell_{ij}=\left\{k:y_{ik}=1,f_{ik}\geq f_{ij}\right\}$ and also $\mathrm{rank}_{ij}=\left|\left\{k:f_{ik}\geq f_{ij}\right\}\right|$.

\subsubsection{Area Under ROC Curve (AUC)}
\label{sssec:subsubhead}
AUC is a threshold-independent metric, defined as the probability that a classifier will rank a randomly chosen positive instance higher than a randomly chosen negative one. An overall accuracy is computed by averaging the AUC score for each instrument.

\subsubsection{$F_1$ Score}
\label{sssec:subsubhead}
$F_1$ is a widely used metric and will prove useful in evaluating our work compared to existing methods, as well as getting insight into possible class imbalance problems. In order to calculate an overall score, we compute the average of the per-instrument scores at both micro and macro scales.\par

\section{Results and Discussion}
\label{sec:majhead}

\textbf{Comparison of Network Architectures:} We first train the models using only the monophonic dataset, in order to get some baseline numbers. The results show that the proposed architecture significantly improves the LRAP score of the model proposed by Gururani et al. \cite{b15} (Table II) and surpasses most of the published efforts according to F1 score (Table IV). We attribute this improvement to the utilization of blocks of consecutive convolutional layers compared to \cite{b15}, as well as the usage of batch normalization in the network compared to \cite{b21}.
Additionally, results produced by a re-implementation of the architecture of \cite{b21} are similar to those originally reported, leading us to assume that the differences in audio preprocessing are of minor importance. 

\textbf{Comparison of Data Augmentation Techniques:} Next, our model is trained using the randomly mixed dataset in order to get an insight into the effects of mixed tracks. As shown in Table~III, the model slightly improves on the monophonic one. In order to confirm that this improvement is attributed to the mixing technique itself and not the increased number of samples, we perform an additional training session on an augmented,
using pitch-shifting of $\pm2,\pm4,\pm6$ semitones, monophonic dataset with an equal number of samples to the mixed one. The outcome does indeed not surpass the randomly mixed augmented dataset, although we observed a slight increase in metrics compared to the initial dataset, including a 77.2\% LRAP.\par

We then proceed to train additional models using the data generated with the three augmentation strategies. A performance comparison between them is summarized in Table~III. The strategy of mixing tracks of the same genre barely improves the result, mixing tempo-synced tracks though scores higher compared to the baseline, with a non-negligible improvement of 1.5\% LRAP compared to random mixing. This indicates that leaving the tempo variations out of context helps the model identify certain instruments. This is also the case for the pitch-synced mixes, which provide 2\% LRAP improvement compared to the random polyphonic baseline and ca. 3\% compared to the monophonic case. 

\begin{table}[]
\vspace{-0.2cm}
 \centering
\caption{Comparison on IRMAS Dataset Performance} \label{tab:sometab}
\begin{tabular}{l||cc}
Models                               &$F_1$micro &$F_1$macro \\ \hline \hline
Bosch et al. {[10]}               & 0.503    & 0.432    \\
Han et al. {[19]}                 & 0.602    & 0.503    \\
Pons et al. {[31]}                & 0.589    & 0.516    \\
Proposed - Baseline                  & 0.616    & 0.506    \\
Proposed - Combined
& \textbf{0.647}    & \textbf{0.546}    \\ 
\end{tabular}

\vspace{-0.5cm}
\end{table}

    
    

    
    
    
 %
    
    

\textbf{Usage of an Ensemble Classifier:} The above results imply that by manipulating a musical attribute such as tempo or pitch, we produce models that identify better certain characteristics. Thus, we expect aggregated predictions of the augmentation models to produce more representative outcomes. 
In order to test this hypothesis, we average the predictions analyzed above for each examined combination of strategies along the same fold, resulting in 5 new sets of predictions for each combination. 
The best performing ensemble model (denoted as Combined) consists of aggregated predictions of all models, excluding the one trained on randomly mixed data. Its performance on the test set (see Table III) yields above 80\% in terms of LRAP, outperforming the monophonic case by almost 4\%, a non-negligible improvement given the complexity of IRMAS test set. Furthermore, the ensemble model compares favorably against a number of recently published experiments on IRMAS, achieving an improvement of approximately 3\% in macro $F_1$-score, and 4\% in micro $F_1$-score. \par
As an additional experiment, we train an ensemble classifier using the procedure described above, but replacing our proposed architecture with the one used by Gururani et al. \cite{b15}. This experimental setup results in a 77.9\% LRAP, further indicating that proper data augmentation can improve instrument recognition results as much as architectural refinements.


\begin{figure*}[t]

\begin{minipage}[t]{1.0\linewidth}
  \centering
  \centerline{\includegraphics[width=\textwidth]{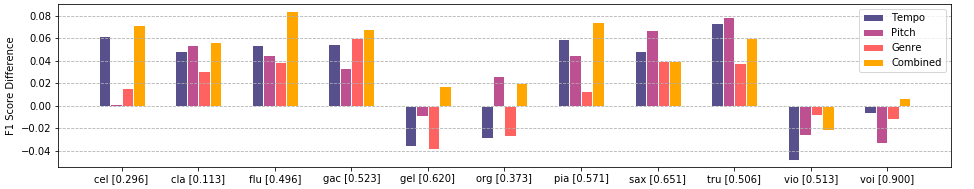}}
\end{minipage}
\caption{Class-wise performance of augmentation models in terms of $F_1$-score difference from the baseline monophonic case, included in brackets.} 
\vspace{-0.5cm}
\end{figure*}  

\textbf{Per-Instrument Results:} Finally, we examine the per-instrument $F_1$ score, in order to get an insight into the effect of specific augmentation methods in recognition of various instrument types. The results are visualized in Fig.~3. We observe that data augmentation mainly improves the recognition results in instruments that have a primarily supporting role or tend to play in unison, such as woodwind instruments. On the other hand, recognition of naturally predominant instruments such as the electric guitar or the violin, and vocals is not improved, or even becomes more difficult, through data mixing. 

\section{Conclusions}
\label{tab:sometab}

In this paper we investigated the problem of using monophonic data to train a polyphonic instrument classifier, focusing on mixing strategies to efficiently convert the monophonic dataset to polyphonic. The proposed model achieves a better interpretation of spectral representations of polyphonic audio, while the proposed mixing techniques proved to be adequate augmentation strategies, outperforming the monophonic case. Tempo and Pitch mixing strategies yield the best results, while the overall best performance is achieved by utilizing an ensemble of classifiers, inheriting different spatio-temporal characteristics from each model. Future work on this task could include mixing techniques focusing on musical traits like key alignment, as well as further investigation into multi-instrumental mixing, since preliminary experiments using random mixes of 3 instruments perform worse than the proposed strategies, indicating that more sophisticated augmentation methods and architectural refinements are required in this case.

\end{document}